\documentclass[journal,10pt]{IEEEtran}

\usepackage[pagewise]{lineno}

\usepackage{hyperref}
\usepackage{verbatim}
\usepackage{graphicx}
\usepackage{multicol}
\usepackage{graphicx}
\usepackage{amsmath}
\usepackage{cuted}
\usepackage{cite}
\usepackage{multirow}
\usepackage{hhline}
\usepackage{xcolor}
\usepackage{balance}
\usepackage{subfigure}
\usepackage{hyperref}

\usepackage{stfloats}

\makeatletter
\def\hlinewd#1{%
\noalign{\ifnum0=`}\fi\hrule \@height #1 \futurelet
\reserved@a\@xhline}

\makeatother

\begin{document}

\title{MCNet: A crowd denstity estimation network based on integrating multi-scale attention module}

\author{Qiang Guo, Rubo Zhang, Di Zhao

\thanks{Qiang Guo, Rubo Zhang are with College of Mechanical and Electronic Engineering, Dalian Minzu University, 116650, Dalian, China.
(e-mail: guoqiang01486@dlnu.edu.cn, zhangrubo@dlnu.edu.cn).}
\thanks{Di Zhao are with School of Computer Science and Engineering, Dalian Minzu University, 116650, Dalian, China, and also with the Dalian University of Technology, Dalian 116024, China and Postdoctoral workstation of Dalian Yongjia Electronic Technology Co., Ltd, Dalian, China.
(e-mail: zhaodi@dlnu.edu.cn).}
}

\maketitle
\begin{abstract}
Aiming at the metro video surveillance system has not been able to effectively solve the metro crowd density estimation problem, a Metro Crowd density estimation Network (called MCNet) is proposed to automatically classify crowd density level of passengers. Firstly, an Integrating Multi-scale Attention (IMA) module is proposed to enhance the ability of the plain classifiers to extract semantic crowd texture features to accommodate to the characteristics of the crowd texture feature. The innovation of the IMA module is to fuse the dilation convolution, multiscale feature extraction and attention mechanism to obtain multi-scale crowd feature activation from a larger receptive field with lower computational cost, and to strengthen the crowds activation state of convolutional features in top layers. Secondly, a novel lightweight crowd texture feature extraction network is proposed, which can directly process video frames and automatically extract texture features for crowd density estimation, while its faster image processing speed and fewer network parameters make it flexible to be deployed on embedded platforms with limited hardware resources. Finally, this paper integrates IMA module and the lightweight crowd texture feature extraction network to construct the MCNet, and validate the feasibility of this network on image classification dataset: Cifar10 and four crowd density datasets: PETS2009, Mall, QUT and SH\_METRO. The experimental results show that, with the help of IMA module, the prediction accuracies of the plain MCNet are improved to varying degrees on these datasets, and demonstrates better overall prediction performance compared to other competitors in accuracy, total number of network parameters and inference speed. Furthermore, the experiments on the power consumption and inference speed of MCNet on a workstation and an embedded device further support the feasibility of deploying MCNet on the embedded metro platform for passengers density estimation, demonstrating that the MCNet can be a suitable solution for crowd density estimation in metro video surveillance where there are image processing challenges such as high density, high occlusion, perspective distortion and limited hardware resources.

\end{abstract}
\section{Introduction}
In the field of metro video surveillance, there has been a demand for using video frames to estimate passengers density. Using this information, the metro video surveillance system can master the crowding in each carriage and intelligently guide waiting passengers to uncrowded carriages to improve their traveling experience, and can also help electrical equipment (e.g. air conditioners) to automatically regulate their operating status to save energy and reduce emission. In addition, abnormal crowd density is often related to abnormal events \cite{Ref1}. Based on accurate crowd density information, metro managers can effectively manage and control metro crowd and prevent unsafe incidents such as stampede. Currently, there are two main methods to achieve crowd density estimation \cite{Ref2}: providing an approximate estimate of the number of crowded people \cite{Ref3,Ref4,Ref4a,Ref4b,Ref4c}, and providing the density level of the crowd \cite{Ref5,Ref6,Ref8,Ref9,Ref6a,Ref6b,Ref10,Ref6c,Ref6d}. The former method (named detection-based method) can roughly estimate the number of the people, but in overcrowding scenes (e.g., metro stations, squares, etc.), is difficult to accurately segment pedestrians for crowd counting in bustling crowd images due to the severe occlusion and perspective distortion of pedestrians \cite{Ref7}, while the latter method (called feature-based method) only needs to give the results of crowd density level classification, which is easier to be realized compared with the former method and more suitable for applications that are not sensitive to the number of people but more concerned about the crowd density level. Therefore, many research work have been carried out in this research field in the past decades.

Davies et al. \cite{Ref5} first proposed to use image processing techniques for crowd density estimation, which uses background suppression and edge detection methods to estimate static crowd density, and then combined them with the optical flow algorithm to calculate crowd movement direction. However, as the density of the crowd increases, its prediction accuracy decreases due to heavy pedestrian occlusion, which renders the edge detection method ineffective and thus unable to accurately estimate the crowd density. Following researchers have proposed different solutions \cite{Ref8,Ref6,Ref9,Ref6a,Ref6b,Ref10,Ref6c,Ref6d}, of which the literature \cite{Ref6d} demonstrated good results in scenes with different crowd densities by utilizing an adaptive crowd density classification method based on pixels and texture features. However, these methods still suffer from the problem that the proposed methods depend on specific scenarios and can't obtain better prediction results when applied across scenarios.

In 2012, the deep CNN AlexNet \cite{Ref11} won the ImageNet \cite{Ref12} competition with a significant margin, which makes CNN become a research hotspot. CNN are artificial implementations of animal visual system, which utilizes local receptive field, spatial information extraction, weights sharing and subsampling to enable it to effectively extract image spatial features. Recently, researchers have proposed crowd density estimation methods based on the deep CNN and achieved good results. Fu et al. \cite{Ref7} proposed to use the CNN to estimate crowd density and enhanced its prediction accuracy by cascading two CNN-based classifiers. Zhang et al. \cite{Ref13} designed a mixed pooling layer to enhance the generalization ability of the CNN, the effectiveness of which was validated on benchmark datasets. Li et al. \cite{Ref14} proposed a novel key-frame extraction technology and then combined it with GoogleNet \cite{Ref15} to build a new crowd density estimation network, and the experimental results showed that the network has better recognition accuracy in benchmark datasets. Huynh et al. \cite{Ref15a} designed a CNN-based encoder-decoder architecture and then adopted the multi-task learning strategies to enhance the feature generality under different scenes, helping the proposed model to estimate crowd density levels accurately. Bhuiyan et al. \cite{Ref16a} proposed a deep crowd dilated convolutional neural network for crowd density analysis to solve their concerns. Pu et al. \cite{Ref16} built a large-scale metro crowd density dataset to test the crowd density estimation performance of deep CNNs, and experimental results show that it is feasible for deep CNNs to accurately estimate the crowd density of metro passengers in real scenarios.

Although the emerging CNN-based crowd density estimation methods have obtained a better prediction accuracy than the tradition methods, but there are still two problem that have not been fully explored and resolved. 1). Distribution characteristics of crowd texture features. The extracted features of image classification problem are the feature activations of a certain part of the target object \cite{Ref17a}. CNN classifies images of different object classes by strongly activating features at specific locations of objects in the image, while weakening features in other regions of the image, so the regular convolutional filter compresses pixels within a fixed area of the image and forms the feature activation on the deep convolutional features. and this localised feature activations are not suitable for dealing with crowd density classification problem. The crowds are dispersed over a larger and wider range in the image, and image regions with different crowd scales and density classes are formed in the image. Therefore, it is necessary that the foreground crowd texture features have a wider activation field and different scales of feature activation, which makes regular convolution operations unable to properly handle the crowd density estimation problem. 2). The computing burden at CNN. The complex matrix operations and hierarchical network structure of CNN make it have high computational resource requirements for the target platform, which hinders the application and arrangement of CNN-based crowd density estimation methods in embedded scenarios. Currently, the main research efforts are still focused on continuously improving the prediction accuracy of CNN-based crowd density estimation methods, ignoring the need to build practical technical solutions under the hardware constraints of embedded environments. Thus it's a more practical research direction to establish a better prediction performance balance between the computational complexity and the prediction accuracy of the network to help the established crowd density estimation network to achieve fast, stable and accurate online prediction in the edge devices with more limited computational resources. To my best of my knowledge, the above two issues are less attention and explored in this research field, motivating me attempt to solve these problems.

\begin{figure*}[]
	\centering
		\includegraphics[scale=.8]{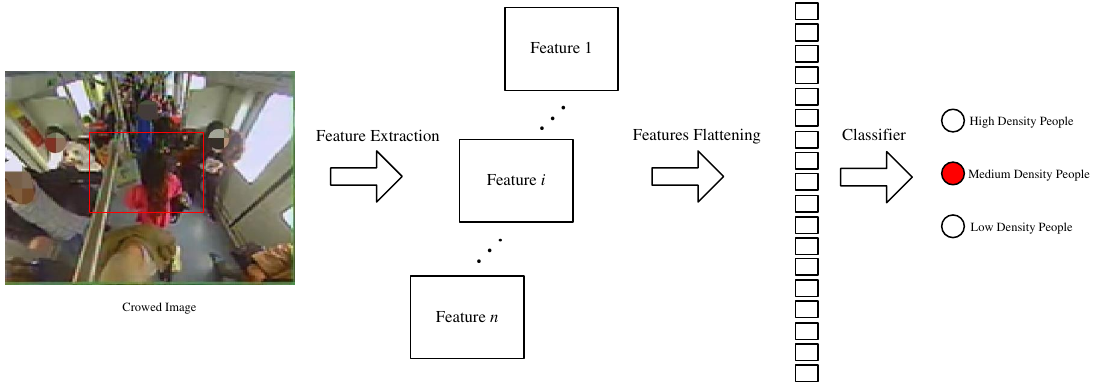}
    \caption{The image processing pipeline of the feature-based crowd density estimation method.}
	\label{FIG:1}
\end{figure*}

This paper utilizes a variety of network structure and architecture designing strategies to build a novel lightweight feature extraction network, which has small computational burden and extract the crowd texture features for automatically classifying passengers density level. Then, in view of the characteristics of crowd texture features, an Integrating Multiscale Attention (IMA) module is proposed by fusing the dilation convolution, multi-scale feature extraction and attention mechanism to enhance crowd texture feature, which helps the plain network to generate crowd texture features with better feature activation distributions. Finally, a MCNet is constructed by combining the IMA module with a lightweight crowd texture feature extraction network, and the feasibility of the network is validated on benchmark datasets and the metro scene dataset, showing that it has the potential to used as a soft instrument to help metro managers track passenger flow on carriages and on platforms. In summary, the contributions of this paper are threefold:


\begin{itemize}
  \item [(1)]
  Aiming at the crowd texture features require a wider ranges of feature activations and different scales of feature activations, an Integrating Multiscale Attention (IMA) module is proposed by combining dilation convolution, multi-scale feature extraction and attention mechanism. Where the dilation convolution enlarges the receptive field of the convolutional filter to cover a larger range of input image pixels at a small computational cost, and multiple dilation convolution with different dilation rate form a multi-branch network structure to achieve multi-scale feature extraction, and then the spatial attention is applied to each network branch to strength crowd features. Thus, these three network design strategies combine together to process the bottom layer features, so that the top layer features have more scales and a wider range of semantic feature activation to strengthen the crowd texture features.
  \item [(2)]
  Combining the lightweight crowd feature extraction network and the IMA module, a lightweight crowd density estimation network (named MCNet) with competitive prediction performance is constructed, which has faster inference speed and fewer number of network parameters and can be deployed on embedded devices with limited resources and provides accurate crowd density estimation results. In order to fully test the feasibility of the IMA module and MCNet, the abliation and benchmark experiments are conducted on five popular datasets, including Cifar10 \cite{Ref28}, PETS2009 \cite{Ref24}, Mall \cite{Ref25} and QUT \cite{Ref26}, and the experimental results of the IMA module can effectively improve the the prediction accuracy of the plain network, and the proposed MCNet has competitive overall prediction performance results compared to other comparative networks, and can be used to accurately estimate crowd density information online on low-performance hardware.
  \item [(3)]
  To validate that the proposed method can solve the challenge of classifying crowded passenger images under constrained computational resources, this paper conducts various experiments to evaluate the ability of the MCNet to solve the crowded passenger density estimation problem in the metro scenes, including the experiment on a large-scale and manually labeled metro crowd density dataset: ShangHai METRO dataset (SH\_METRO) and inference speed and power consumption experiments. The experimental results show that MCNet can be used as a solution to the problem of metro crowd density estimation, and can be used as a soft instrument to replace manual monitoring and provide metro operators with real-time, accurate information on metro crowd density.
\end{itemize}

The rest of this paper is structured as follows. Section \ref{sec:2} introduces the methodology of the proposed method, where the theory of the IMA module to strength crowd features and the design strategies of the MCNet are elaborately explained. The experiment results are presented and discussed in Section \ref{sec:3}. Finally, Section \ref{sec:4} is the conclusion.
\section{Methodology}\label{sec:2}
In this section, this paper detailed explains the theory of the feature-based crowd density estimation method and designing strategies of the lightweight crowd texture feature extraction network, and then theory of the IMA module and the modeling approach of MCNet are elaborated.

\subsection{The theory of the feature-based crowd density estimation method}
The feature-based crowd density estimation method can effectively predict the crowd density level. In order to clearly explain this method, the image processing pipeline of which is drawn in Fig.~\ref{FIG:1}. For a crowd image with a medium density level, the feature-based crowd density estimation method first extracts features from the crowd image, and then processes the extracted features and feeds them to the top classification network, which learns to use these features to estimate crowd density using a supervised algorithm. Thus feature extraction is a crucial procedure of the feature-based crowd density estimation method and the extracted features are called texture features. Texture features are an important image features and can be used for a variety of image processing tasks. Since texture features can express the density distribution in an image, they are ideally suited to solve the crowd density estimation problem. In crowd images, highly crowded people tend to exhibit strong texture feature activation and the feature activation gradually decrease as the crowd density level of the image decreases \cite{Ref1}.

Previous researchers have utilized structure and statistical methods to extract image texture features, but the extracted features have limited expressiveness and can't effectively distinguish between crowd and background in complex scenes. Therefore this paper aims to use Convolutional Neural Network (CNN), which has good feature extraction and representation capability, to solve the problem of image texture feature extraction, and the principles are as follows.

The CNN is a multi-layer feed-forward neural network. Assuming that the network has $k$ layers, its architecture can be represented as:

\begin{eqnarray}
{\phi }_{k}}\circ \cdots \circ {{\phi }_{2}}\circ {{\phi }_{1}
\end{eqnarray}
where $\circ$ denotes layer connections, and ${{\phi }_{i}},i=1,2,\cdots ,k$ stands for a network layer.

For a input features $\textbf{\emph{X}}\in {\textbf{\emph{R}}^{H\times W\times C}}$, the feature activation of the network is:

\begin{eqnarray}
\textbf{\emph{Y}}=({{\phi }_{k}}\circ \cdots \circ {{\phi }_{2}}\circ {{\phi }_{1}})(\textbf{\emph{X}})
\end{eqnarray}

$\textbf{\emph{Y}}\in {\textbf{\emph{R}}^{{H}'\times {W}'\times {C}'}}$ is considered as the feature description matrix obtained from the input image after the feature extraction operation. Given that the network is only composed of convolutional and pooling layers, the equation (2) can be rewritten as:

\begin{eqnarray}
\textbf{\emph{Y}}={\textbf{\emph{W}}_{k}}\left\{ \cdots \{{\textbf{\emph{W}}_{2}}*[({\textbf{\emph{W}}_{1}}*\textbf{\emph{X}})\downarrow {\textbf{P}_{1}}]\}\downarrow {\textbf{P}_{2}}\cdots  \right\}\downarrow {\textbf{P}_{k}}
\end{eqnarray}
where $*$ denotes the convolutional operation and $\downarrow$ represents pooling operation. ${\textbf{\emph{W}}_{i}}$ is convolutional filter matrix and $\textbf{P}_{i}$ is the sampling matrix. For simplicity of formulation, CNN activation function and biases are omitted.

During the training process, there are different combinations of values of the convolutional filter matrix ${\textbf{\emph{W}}_{i}}$ to extract different image features. For example, when the sum of all elements in the convolutional filter matrix ${\textbf{\emph{W}}_{i}}$ is zero, it achieves the function for extracting the corner features of the input image, which are very important features in texture analysis \cite{Ref20}. According to equation (3), the features extracted by a layer will continue to propagate backward, and the following network layers will perform various nonlinear combinations of these features to learn to express the texture features of the input image.

The CNN has a large number of network parameters, which can store a large amount of texture features. Meanwhile, its spatially longitudinally extended network structure makes these texture features establish a non-linear mapping relationship between the crowd density information contained in the image and the crowd density level. Based on this principle, this paper proposes a CNN-based crowd density estimation network to efficiently extract texture features from crowd images at lower computational cost, and then uses the extracted texture features to automatically predict the crowd density level of these images, the designing strategies and architecture of which will be explained in the next section.

\subsection{The designing strategies of the lightweight crowd texture feature extraction network}
CNN is mainly composed of convolutional layers and fully connected layers. The total computation of the network is occupied mostly by the inference operations of each convolutional layer, while the total number of network parameters is mainly occupied by the weight parameters of each fully connected layer \cite{Ref21}. Therefore, all layers of the network chooses the convolutional layer to reduce total number of network parameters, and the number of convolutional filters is decreased to lower computing complexity. In the choice of filter size, this paper uses the $1\times1$ convolutional filters to merge convolutional features in different channel and $3\times3$ convolutional filters to extract texture features, and enables the network to have better feature extraction ability as lower computational cost by stacking the two filters. Ma et al. \cite{Ref19} has pointed out that the multi-branch CNN slows down its inference speed, thus the preferred architecture of the proposed network is single column architecture. Meanwhile, in order to enrich the crowd passengers features, the paper utilizes the refined multi-branch convolutional block: fire module to obtain multi-scale passengers features on the top layers. Finally, the proposed network also needs to have enough receptive field to ensure that the top layers cover the whole input image. The receptive field is computed as follows \cite{Ref22}.

\begin{eqnarray}
{{r}_{n}}={{r}_{n-1}}+({f}_{n}-1)\prod\limits_{i=1}^{n-1}{{{s}_{i}}}
\end{eqnarray}
where $r_n, r_{n-1}$ stand for the receptive field of $n$th layer and $(n-1)$th layer respectively. $f_n$ represents the filter size of $n$th layer. $s_i$ represents the stride of $i$th layer.

\begin{figure}[!h]
	\centering
		\includegraphics[width=0.35\textwidth]{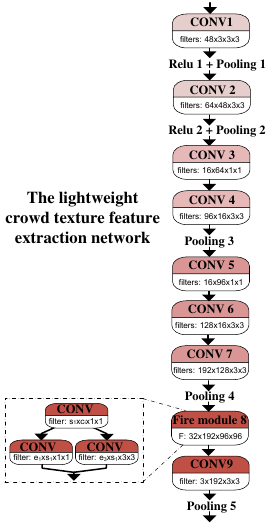}
    \caption{The architecture of the lightweight crowd texture feature extraction network, where ``CONV'' represents the convolutional layer and ``filter'' denotes convolutional filter with the shape of output channels $\times$ input channels $\times$ filter height $\times$ filter width. ``F'' represents the filters shape of the fire module. ``Pooling'' represents the pooling layer. The ReLU layers after the convolutional layers are omitted for clear presentation.}
	\label{FIG:2}
\end{figure}

\begin{figure*}[]
	\centering
		\includegraphics[width=0.6\textwidth]{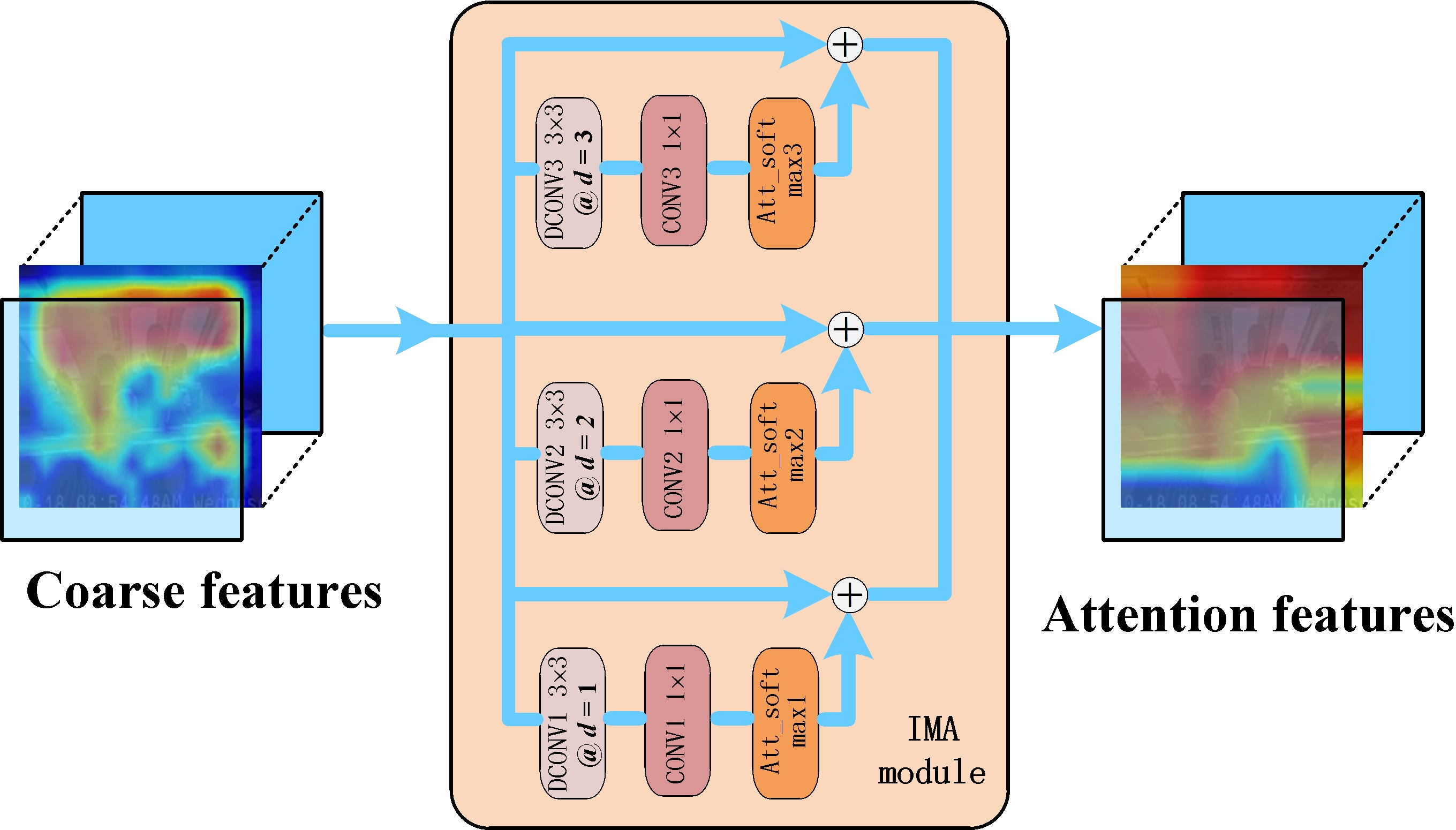}
    \caption{The structure of the IMA module, where ``DCONV1-3'' and ``CONV1-3'' represent the dilation convolutional layer and convolutional layer of different branches. ``$3\times3$'' and ``$1\times1$'' denote their filter shape. ``\emph{d}'' represents the dilation rate. ``Att\_softmax1-3" denote softmax layer of different branches to realize gate mechanism.}
	\label{FIG:3}
\end{figure*}

Using the equation (4), the receptive field of 15-layers network is 374 (its structure is draw in Fig.~\ref{FIG:2}, which can fully cover an input image with a shape of $227 \times 227$. Considering that the actual receptive field of the network during its inference process is often lower than this theoretical value \cite{Ref23}, a larger receptive field can help the network to extract global features effectively. In summary, the following several network designing strategies are proposed in this paper to reduce the computational burden of the network while maintaining good network prediction accuracy.

\begin{itemize}
  \item [(1)]
  The single column network architecture is adopted to keep its inference speed.
  \item [(2)]
  All layers are convolutional layer to reduce total number of network parameters.
  \item [(3)]
  The preferred shape of convolutional filter is $1\times1$ and $3\times3$ to limit the computing complexity.
  \item [(4)]
  The network structure of 15-layers CNN ensures it has enough receptive filed to cover input image.
  \item [(5)]
  The fire module convolutional block is adopted in top layer of the network to extract multi-scale passenger features.
\end{itemize}
where strategies (1),(2) and (3) are used to decrease its computational burden, and strategy (4) and (5) are utilized to ensure network prediction accuracy. This paper utilizes these strategies to construct the overall network structure and help it achieve good feature extraction and crowd density estimation capability with a small model size, the architecture of which is shown in Fig.~\ref{FIG:2}.

As is shown in the Fig.~\ref{FIG:2}, the network has shallow and compact network structure, which matches the hardware constraints of metro embedded platform and can be deployed on it. The network first receives the real-life metro crowd image and processes it over the feature extraction and classification operation, and then outputs the real-time crowd density estimation results. 

\subsection{The IMA module}
\begin{figure*}[]
	\centering
		\includegraphics[scale=.65]{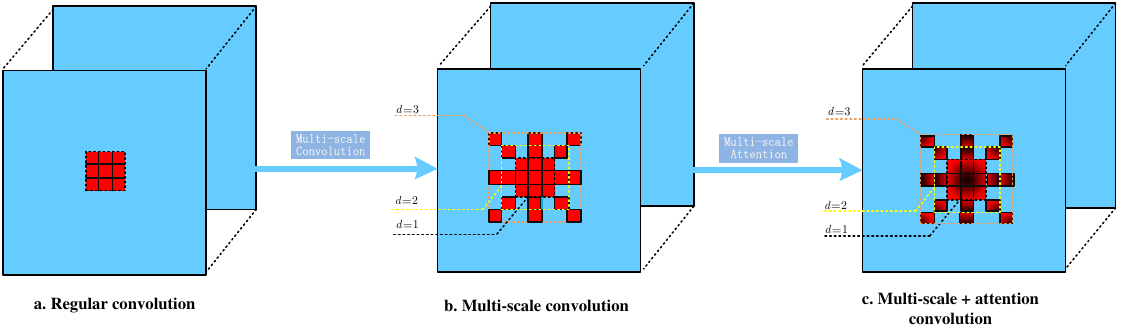}
    \caption{Illustration of the mechanism of the IMA module. The red rectangle in the subgraph a represents a convolutional filter with filter size of ``$3\times3$''. In the subgraph b, the dotted lines with different color denote the receptive field of the different dilation convolutional layers. The red rectangles with black shading in the subgraph c means that attention mechanism has been incorporated to strengthen crowd features.}
	\label{FIG:4}
\end{figure*}

As is mentioned in the above, the CNN is weaken in obtaining wider ranges and different scales of crowd feature activations. To tackle this problem, this paper constructs the IMA module that can fuses the dilation convolution, multi-scale feature extraction and attention mechanism to obtain multi-scale crowd feature in a larger receptive field, the structure of which is shown in Fig.~\ref{FIG:3}. As can be seen that the IMA module uses the convolution operation on each network branch to realize the attention mechanism to fuse and strengthen the regional features, while uses the dilation convolution and multi-branch network structure to adjust the effect field of the attention mechanism to obtain different scales of enhance features. The two work together to help the plain network improve the classification accuracy of crowd density estimation. Fig.~\ref{FIG:4} illustrates the mechanism of the IMA module on the feature map.

For a input feature $\textbf{\emph{X}}$, a regular convolutional layer with filter size $f$ processes it and outputs the feature map $\textbf{X}_o$, which can be represented.

\begin{eqnarray}
\textbf{\emph{X}}_o=\textbf{\emph{W}}\textbf{\emph{X}}
\end{eqnarray}
where $\textbf{\emph{W}}$ can extract crowd features in $f$ size region. In order to expand cover region of the convolutional layer on input features, the dilation convolutional layer is used to replace this convolutional layer, as follow.

\begin{eqnarray}
\textbf{\emph{X}}_\text{o}^1=\textbf{\emph{W}}_d\textbf{\emph{X}}
\end{eqnarray}
where $\textbf{\emph{X}}_\text{o}^1$ represents the output feature maps processed by the dilation convolution layer. $\textbf{\emph{W}}_d$ denotes the its filter matrix. The corresponding filter size $f_d$ is expanded by following equation.

\begin{eqnarray}
f_d=f+(f-1)(d-1)
\end{eqnarray}
where $d$ is dilation rate, which can enlarge the convolutional filter size. thus this paper extracts crowd features at different scales by setting different dilation rates in different network branches, and then concatenates them to obtain multi-scale features, the mechanism of which is shown in Fig. 4b.

We further process Equation 4 to strength the crowd feature activation. Since the channels of the feature maps contain spatial feature activation information of the different crowd density level during the training process, in order to utilise this information, the channel-wise convolutional operation (convolutional filter size is $1\times1$) is used to merge channel features.

\begin{eqnarray}
\textbf{\emph{X}}_\text{o}^2=\textbf{\emph{W}}_m\textbf{\emph{W}}_d\textbf{\emph{X}}
\end{eqnarray}
where $\textbf{\emph{W}}_m$ denotes the filter matrix of $1\times1$ convolutional layer. $\textbf{\emph{X}}_\text{o}^2$ represents the merged crowd features.

Through two-stage convolutional operations, $\textbf{\emph{X}}_\text{o}^2$ already contains a certain amount of global spatial information. In order to use these information, a gating mechanism (a softmax function $\lambda$ ) is used to obtain the enhanced crowd features.

\begin{eqnarray}
{\textbf{\emph{{X}}}_\text{att}} = \lambda(\textbf{\emph{X}}_\text{o}^2)
\end{eqnarray}

Then, the $\textbf{\emph{{X}}}_\text{att}$ are merged into the each position of input features $\textbf{\emph{X}}$ to aggregate the global texture features.

\begin{eqnarray}
{\textbf{\emph{{X}}}_o^\text{att}} = {\textbf{\emph{{X}}}_\text{att}} + {\textbf{\emph{{X}}}}
\end{eqnarray}

The mechanism of the attention mechanism is shown in Fig. 4c.

\begin{figure}[!h]
	\centering
		\includegraphics[scale=.7]{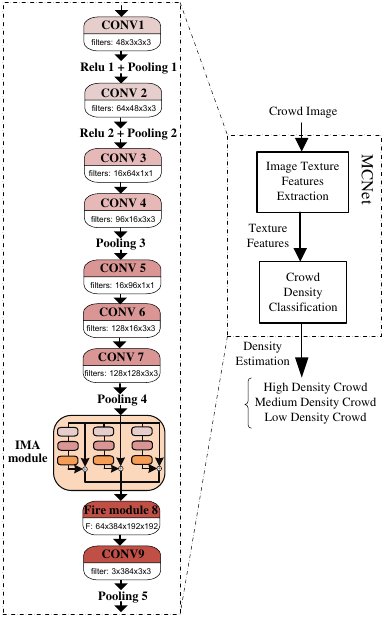}
    \caption{The architecture of the MCNet.}
	\label{FIG:5}
\end{figure}

\begin{table*}[bp]\label{alg1}
\centering\hypertarget{alg1}{}
\begin{tabular}{l}
\hlinewd{1pt}
\textbf{Algorithm 1} Image processing procedures of the MCNet                                                                                                                                                                                                                                                                                                                                                                                                                                                                                                                                 \\ \hline
\begin{tabular}[c]{@{}l@{}}\textbf{Input}: $\textbf{X}$, which represents a metro crowd image.\\ \textbf{Output}: $y$, which denotes the density level of the metro crowd image.\\ \textbf{Target}: $f: \textbf{X} \to y$, which means the functional relationship between \textbf{Input} and \textbf{Output}.\\ 1. Obtain meaningful semantic features $\textbf{X}_{\rm{texture}}$ by processing $\textbf{X}$ over the crowd texture feature \\ extraction network.\\ 2. Obtain multi-scale attention crowd features $\textbf{X}_{\rm{IMA}}$ by processing $\textbf{X}_{\rm{texture}}$ over the IMA module. \\ 3. Obtain flattened features $\emph{\textbf{y}}_{\rm{texture}}$ by processing $\textbf{X}_{\rm{IMA}}$ over the classification subnetwork.\\ 4. Obtain $y$ by processing $\emph{\textbf{y}}_{\rm{texture}}$ with a softmax function.\\ \hline\end{tabular} \\ 
\hlinewd{1pt}
\end{tabular}
\end{table*}

\subsection{The modeling method of the MCNet}
This paper analyses the historical surveillance videos of metro station and find that the crowd density levels of metro passengers can be broadly classified into three categories: high, medium and low, so the proposed network requires to have the capability to classify metro passengers into three crowd density levels and helps the metro video surveillance system collect passenger density information to assist the surveillance of the metro station. For example, this network can estimate passengers flow at each metro door and display three different coloured (red, yellow, green) lights at the top of the door to alert waiting passengers to the crowd density level in this carriage. For crowded doors, the light will displays a red or yellow light (depending on the crowd density level), as well as the carriage number of the uncrowded doors, intelligently guiding passengers to the uncrowded doors (showing a green light) to wait for the metro, providing a better travelling service for passengers. To satisfy these requirements, the crowd texture feature extraction network is used as the backbone of MCNet, which has better feature extraction capability and small model size, and can be used to solve the problem of crowd density estimation in metro scenarios with limited hardware resources. Then, the IMA module is used to enhance the activation state of crowd features and improve its distribution characteristics, so that the top classifier of MCNet can make use of these high-quality features to better capture the information of crowd density level from the input image, and promote its the classification accuracy. The architecture of the MCNet is shown Fig.~\ref{FIG:5} and its detailed procedures to process a metro crowd image is listed in Algorithm 1.

\section{Experiments and results}\label{sec:3}
\subsection{Datasets}
\begin{figure}[]
	\centering
		\includegraphics[width=0.45\textwidth]{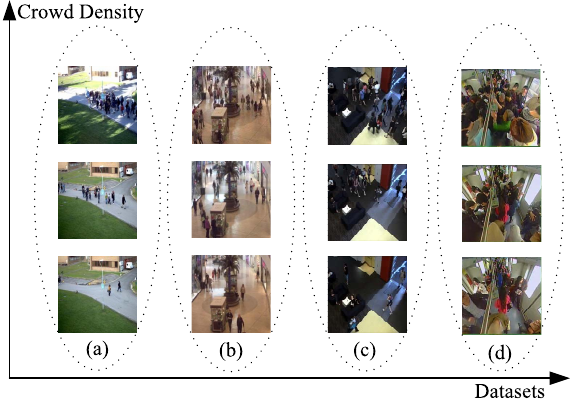}
    \caption{Crowd density distribution: (a) PETS2009; (b) Mall; (c) QUT; (d) SH\_METRO.}
	\label{FIG:6}
\end{figure}

The benchmark and SH\_METRO datasets are used to test the crowd density estimation performance of the MCNet in the following experiments. The benchmark datasets include Cifar10, PETS2009, Mall and QUT datasets. The Cifar10 dataset is popular image classification benchmark dataset and can be used to validate the general capability of the IMA module in solving image classification problem. The images in other datasets are collected from various urban life scenes and have been widely used by researchers in crowd density estimation. In addition, since a public and large-scale metro crowd density dataset is not available, this paper establishes a metro crowd density dataset: SH\_METRO for metro crowd density estimation experiments, which is collected from an metro line in Shanghai, China, and contains a total of 3834 images from various metro lines at different times. The crowd density distribution of the crowd denstity datasets have been shown in Fig.~\ref{FIG:6} according to the population distribution in Table~\ref{Tbl1}, and the experiments were conducted according to the samples distribution of training and testing dataset in Table~\ref{Tbl2}.

\begin{table}[!h]
\centering
\caption{Population distribution under three crowd densities.}\label{Tbl1}
\setlength{\tabcolsep}{0.2em}
\begin{tabular}{llll}
\hline
Dataset               & High Density                 & Medium Density             & Low Density             \\ \hline
PETS2009              & \textgreater{}19             & 10-19                      & 0-9                     \\
Mall                  & \textgreater{}19             & 10-19                      & 0-9                     \\
QUT                   & \textgreater{}19             & 10-19                      & 0-9                     \\
SH\_METRO             & \textgreater{}12             & 7-12                       & 0-6                     \\ \hline
\multicolumn{4}{l}{\begin{tabular}[c]{@{}l@{}}{}\end{tabular}}
\end{tabular}
\end{table}

\begin{table}[!h]
\centering
\caption{Samples distribution of the crowd denstity datasets.}\label{Tbl2}
\setlength{\tabcolsep}{0.2em}
\begin{tabular}{llllllll}
\hline
\multirow{2}{*}{Dataset}       & \multicolumn{3}{l}{Training set}                   & \multicolumn{3}{l}{Testing set}                 & \multirow{2}{*}{Sum.}      \\ \cline{2-7}
                               & H.              & M.              & L.             & H.             & M.             & L.            &                            \\ \hline
PETS2009                       & 640             & 635             & 648            & 315            & 279            & 293           & 2810                       \\
Mall                           & 250             & 293             & 278            & 106            & 122            & 120           & 1169                       \\
QUT                            & 536             & 1049            & 1013           & 536            & 1049           & 1014          & 5197                       \\
SH\_METRO                      & 1473            & 345             & 772            & 685            & 182            & 377           & 3834                       \\ \hline
\multicolumn{8}{l}{\begin{tabular}[c]{@{}l@{}}{\scriptsize ``H.",``M." and ``L." represent the high, medium and low crowd} \\ {\scriptsize density respectively. ``Sum." denotes the total number of samples} \\ {\scriptsize in these datasets}.\end{tabular}}
\end{tabular}
\end{table}

\subsection{System Setting}
\emph{Data preprocessing}. In order to make the network converge easily during the training process, the images in these datasets are normalized to [0, 1] and the shape of the images are resized to 227$\times$227, while the shuffle operation is performed on these datasets.

\emph{Experimental platform}. The experiments have been done on the two experimental platforms: workstation and embedded platform, the hardware and software information of the which are listed in Table~\ref{Tbl3}. The workstation is based on x64 architecture and configured with two graphics cards to support the training and testing of the proposed models. The workstation is a conventional commercial high-performance computing platform, which is widely used in the field of deep learning, and has a larger subscriber base and sufficient references. Therefore this paper does't give a detailed description of this workstation, but only presents its hardware and software, instead focusing on the embedded development board used in the experiments.

The embedded development board is based on Rockchip's RK3399 system-on-a-chip solution, which is big.LITTLE-based CPU designs, integrating dual-core Cortex-A72, quad-core Cortex-A53 and ARM-T860 GPU, providing powerful computing capabilities while optimizing floating-point and fixed-point computing and power consumption. Moreover, it natively supports Android and Ubuntu, which promotes the ease of use of various deep learning frameworks.

The chosen embedded development board can be applied in the metro embedded environment, so the experiments on this device can fully validate the feasibility of the MCNet on the metro scenes with limited hardware resources. In addition, a highly efficient and light deep learning framework: MNN\footnote{https://github.com/alibaba/MNN} is used to accelerate the inference of all models in both two platforms.

\begin{table}[]
\centering
\caption{The detail hardware and software information of experiment platforms}\label{Tbl3}
\setlength{\tabcolsep}{0.2em}
\begin{tabular}{lll}
\hline
\multirow{2}{*}{\textbf{\begin{tabular}[c]{@{}l@{}}Hardware \& \\ Software\end{tabular}}} & \multicolumn{2}{c}{\textbf{Platforms}}                \\ \cline{2-3}
                                                                                          & Workstation           & RK3399PRO   \\ \hline
CPU                                                                                       & Intel Core i7-6700k   & ARM Cortex-A72 \& A53 \\
MEMORY                                                                                    & 64G                   & 4G                           \\
GPU                                                                                       & NVIDIA Geforce GTX 1080 & ARM Mali-T860                \\
Operating System                                                                          & Ubuntu 16.04          & Debian 9                     \\ \hline
\end{tabular}
\end{table}

\emph{Training \& Testing metrics}. The crowd images are used as training input to the network and the output of network are 0,1,2, representing the low, medium and high crowd density respectively. The inference time, total number of network parameters and prediction accuracy are used as the evaluation metrics to reflects all model's comprehensive performance.

\subsection{Baseline models}
In this paper, several popular deep CNNs: AlexNet, GoogleNet, ResNet \cite{Ref27}, SqueezeNet \cite{Ref17} and MobileNet \cite{Ref18} are used to compare with the MCNet, which can demonstrate the superiority of the proposed network that achieve competitive prediction accuracy with less computing and memory resources.

\begin{table*}[]
\centering
\caption{Results of Cifar10 from SqueezeNet using the IMA module}\label{Tbl4}
\setlength{\tabcolsep}{0.2em}
\begin{tabular}{ccccc}
\hline
\textbf{\begin{tabular}[c]{@{}l@{}}Classifier @ \\ Cifar 10\end{tabular}} & MACs (G) & Params(M) & $\rm{Speed^*_{CPU}}$ &Accuracy(\%)    \\ \hline
SqueezeNet                                                                 & 0.288(+0.084)   & 0.728(+0.43) & 5(+1)      & 74(+3) \\
\hline
\multicolumn{5}{l}{\begin{tabular}[c]{@{}l@{}}{\scriptsize ``*" means the experimental results are obtained using the MNN inference engine, similarly hereinafter.}\end{tabular}}
\end{tabular}
\end{table*}

\begin{table*}[]
\centering
\caption{The performance results of the MCNet and other competitors on the benchmark datasets.}\label{Tbl5}
\setlength{\tabcolsep}{0.2em}
\begin{tabular}{lllllllll}
\hline
\multirow{2}{*}{\textbf{Models}} & \multirow{2}{*}{\begin{tabular}[c]{@{}l@{}}MACs\\    (G)\end{tabular}} & \multirow{2}{*}{\begin{tabular}[c]{@{}l@{}}Params\\    (M)\end{tabular}} & \multicolumn{2}{l}{\textbf{PETS2009}}                                                                                               & \multicolumn{2}{l}{\textbf{Mall}}                                                                                                   & \multicolumn{2}{l}{\textbf{QUT}}                                                                                                    \\ \cline{4-9}
                        &                                                                           &                                                                             & \begin{tabular}[c]{@{}l@{}}Accuracy\\   (\%)\end{tabular} & \begin{tabular}[c]{@{}l@{}}$\rm{Speed_{CPU}^{*}}$\\    (ms/frame)\end{tabular} & \begin{tabular}[c]{@{}l@{}}Accuracy\\    (\%)\end{tabular} & \begin{tabular}[c]{@{}l@{}}$\rm{Speed_{CPU}}$\\    (ms/frame)\end{tabular} & \begin{tabular}[c]{@{}l@{}}Accuracy\\    (\%)\end{tabular} & \begin{tabular}[c]{@{}l@{}}$\rm{Speed_{CPU}}$\\    (ms/frame)\end{tabular} \\ \hline
AlexNet                 & 0.724                                                                     & 61                                                                          & 92                                                            & 17                                                                 & 75                                                            & 17                                                                 & 92                                                            & 16                                                                 \\
GoogleNet               & 1.604                                                                     & 7                                                                           & 91                                                            & 20                                                                 & 80                                                            & 21                                                                 & 93                                                            & 20                                                                 \\
ResNet34                & 3.68                                                                      & 25.5                                                                        & 94                                                            & 33                                                                 & 75                                                            & 33                                                                 & 94                                                            & 34                                                                 \\
MobileNet               & 0.569                                                                     & 4.2                                                                         & 92                                                            & 7                                                                 & 70                                                            & 6                                                                 & 92                                                            & 6                                                                 \\
SqueezeNet                  & 0.288                                                                     & 0.7                                                                         & 93                                                            & 5                                                                  & 78                                                            & 5                                                                  & 92                                                            & 4                                                                  \\
\textbf{MCNet w/o}                  & 0.125                                                                     & 0.13                                                                         & 93                                                            & 2                                                                  & 78                                                            & 2                                                                  & 92                                                            & 2     \\
\textbf{MCNet w/}                  & 0.153                                                                     & 0.72                                                                         & 95                                                            & 3                                                                  & 82                                                            & 3                                                                  & 94                                                            & 3     \\\hline
\end{tabular}
\end{table*}

\subsection{The ablation experiment}
This paper conducts the ablation experiment on Cifar10 dataset to test the effectiveness of the IMA module to promote the classification accuracy of the plain network on the general image classification problem. The SqueezeNet, a network with smaller model size and better classification accuracy, is chosen as the plain classifier to validate the general capability of the IMA module to promote the image classification accuracy of the small-scale network. The experiment results are listed in Table~\ref{Tbl4}, wherein ``MACs" represents the Multiply-ACcumulate operations (``G" means billion, lower is better), which measures the computing complexity of the network but it can't replace the actual speed of the network running on the hardware device. ``Params" represents the total number of network parameters (``M" means million, lower is better). ``$\rm{Speed_{CPU}}$" denotes inference time per a video frame on the CPU mode, and the unit of measurement is milliseconds/frame (ms/frame, lower is better). Besides, the data in brackets means the variations in experimental results of the plain network aided by the IMA module.

As can be seen in Table 4, with the help of the IMA module, the accuracy of the SqueezeNet has been increased by 3\% with a relatively small increase in additional computational cost, including 0.084G increase in MACs, 0.43M increase in network parameters, and 1 second increase in inference time. These computational costs come from the feature augmentation operations of the IMA module by inserting it into the SqueezeNet, whose smaller additional computing overheads don't diminish the applicability of the network to the embedded platform, but help it achieve a comprehensive prediction performance. This suggests that the IMA module is effective in enhancing the image classification accuracy of the plain network, which provides multi-scale attention features that can strengthen the object activation state with larger region in the image.

\subsection{Benchmark experiments}
The three benchmark datasets are utilized to validate the feasibility of the IMA module to improve the accuracy of the plain network to estimate crowd denstity and the proposed MCNet whether has competitive overall prediction performance on crowd density estimation problem compared with other mainstream networks. The experiments on three benchmark datasets are conducted and the detailed experimental results are listed in the Table~\ref{Tbl5}, where the \textbf{MCNet w/o} and \textbf{MCNet w/} indicate that it is available without or with the IMA module. The data in this table show that:

\begin{itemize}
  \item [1)]
  The proposed MCNet achieves a competitive crowd denstity estimation performance, which has higher accuracy of 92\%, small amount of network parameters of 0.13M and fast inference speed of 3ms/frame on the QUT dataset (similar experimental results on other datasets), which indicates that proposed structure and architecture methods for CNN can effectively extract the texture features of the crowd image and process them to output crowd density estimation results. The established IMA module further improves the accuracy of the MCNet by 2 and 4\% because it makes these extracted crowd features has stronger activation state in larger range in the deeper layers and help the top classifer of the MCNet to use these high-level features to estimate the density level of the crowd images more accurately.
  \item [2)]
  Compared to recognition performance of the MCNet and the other deep CNNs, although these baseline models have demonstrated better recognition accuracy, they also have drawbacks in huge number of network parameters and slower inference speed, which make these networks are not suitable for being deployed in the embedded platform. Moreover, comparing to the MobileNet and SqueezeNet that are specifically oriented to the embedded platform, and the MCNet demonstrates better overall performance in crowd density estimation considering in these three metrics, especially with the help of the IMA module, the comprehensive prediction performance of MCNet is further strengthened. Thus, The MCNet is more suitable for embedded environment, especially for FPGA device, because the total parameters of the MobileNet are still too huger compared to the proposed network.
  \item [3)]
  The recognition accuracy of all networks on the PETS2009 and QUT datasets are significantly better than that on the Mall dataset, which is mainly due to the sparser crowd distribution of the Mall dataset, unlike PETS2009 and QUT dataset have more apparent crowded characteristics in images (refer to Fig.~\ref{FIG:6}).
\end{itemize}

\begin{table}[]
\centering
\caption{The performance results of the MCNet and the competitors on the SH\_METRO dataset.}\label{Tbl6}
\setlength{\tabcolsep}{0.2em}
\begin{tabular}{lllll}
\hline
\multirow{2}{*}{\textbf{Models}} & \multirow{2}{*}{\begin{tabular}[c]{@{}l@{}}MACs\\ (G)\end{tabular}} & \multirow{2}{*}{\begin{tabular}[c]{@{}l@{}}Params   \\ (M)\end{tabular}} & \multicolumn{2}{l}{\textbf{SH\_METRO}}                                                                                      \\ \cline{4-5}
                               &                                                                     &                                                                          & \begin{tabular}[c]{@{}l@{}}Accuracy  \\ (\%)\end{tabular} & \begin{tabular}[c]{@{}l@{}}$\rm{Speed_{CPU}}$  \\ (ms/frame)\end{tabular} \\ \hline
AlexNet                        & 0.724                                                               & 61                                                                       & 96                                                        & 17                                                             \\
GoogleNet                      & 1.604                                                               & 7                                                                        & 96                                                        & 20                                                             \\
ResNet34                       & 3.68                                                                & 25.5                                                                     & 94                                                        & 33                                                             \\
MobileNet                      & 0.569                                                               & 4.2                                                                      & 96                                                        & 6                                                             \\
SqueezeNet                         & 0.288                                                               & 0.7                                                                      & 95                                                        & 4                                                              \\
\textbf{MCNet w/o}                         & 0.125                                                               & 0.13                                                                      & 93                                                        & 2                                                              \\
\textbf{MCNet w/}                         & 0.153                                                               & 0.72                                                                      & 95                                                        & 3                                                              \\\hline
\end{tabular}
\end{table}

\begin{figure}[]
\centering
\subfigure[High density]{
\includegraphics[scale=.35]{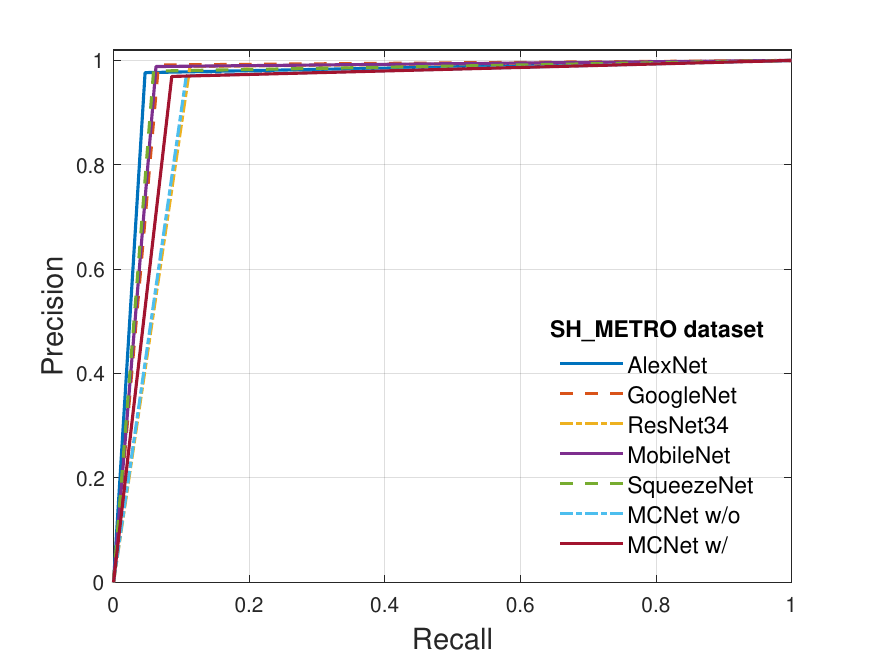}
}
\subfigure[Medium density]{
\includegraphics[scale=.35]{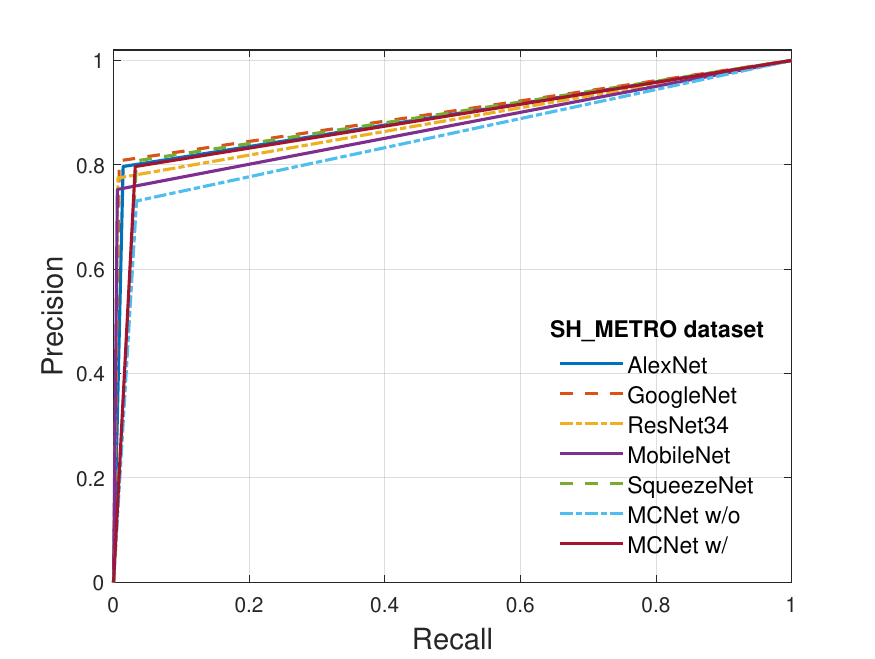}
}
\subfigure[Low density]{
\includegraphics[scale=.35]{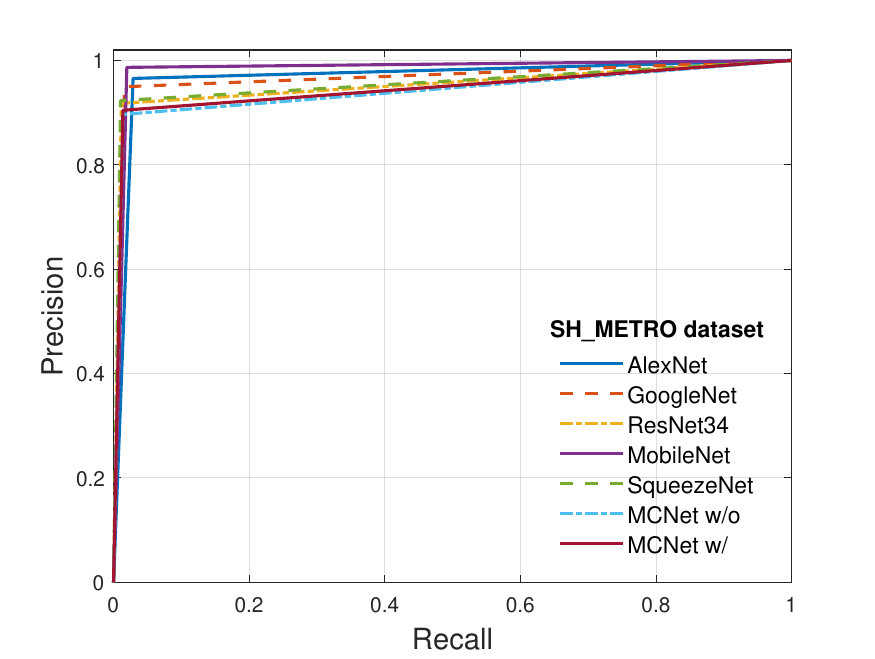}
}
\caption{Precision recall curves of all networks on the SH\_METRO dataset. (a) High density. (b) Medium density. (c) Low density.}
\label{FIG:7}
\end{figure}

\begin{figure*}[bp]
	\centering
		\includegraphics[width=1.0\textwidth]{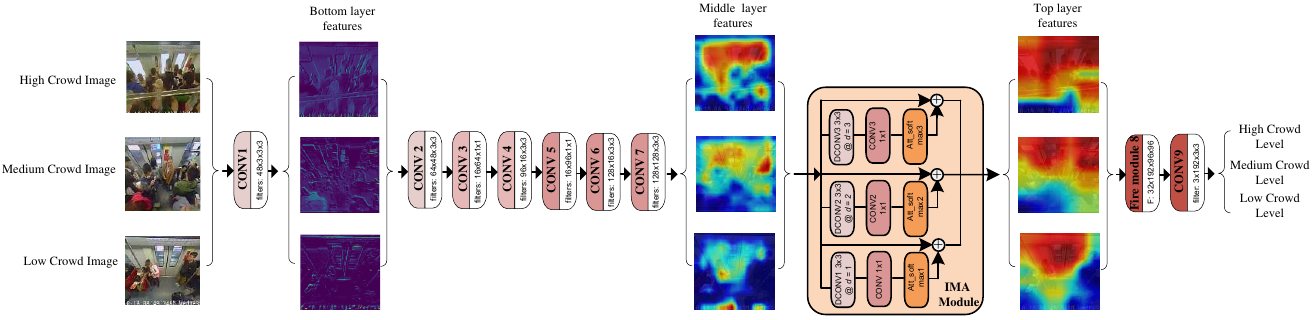}
    \caption{Metro crowd features are generated in different layers of the MCNet with the IMA module.}
	\label{FIG:8}
\end{figure*}

\subsection{The SH\_METRO experiment}
To evaluate the effectiveness of the MCNet's ability to accurately estimate passenger density in real-life metro carriages and the percentage increase in prediction accuracy when equipped with the IMA module, experiments are conducted on the SH\_METRO dataset and the detailed experimental results are shown in Table~\ref{Tbl6}. The recognition accuracy of the MCNet is 93\% and its inference time is only 2ms, indicating that it has a fast image processing speed, which is an important performance metric for the high-speed metro. Moreover, its total parameters is only 0.13M, which can greatly reduce the memory footprint of the target platform. Aided by the IMA module, the prediction accuracy of the MCNet is increased by 2\% with the total network parameters and inference speed increase by 0.59M and 1ms, respectively. Despite witnessing a certain amount of additional computational cost, MCNet achieves higher prediction accuracies at the same computational cost compared to other small models, and thus its overall prediction performance remains competitive and it can be a practical solution for the problem of metro crowd density estimation when considering the evaluation metrics of accuracy, total number of network parameters and inference time together, which is more suitable for deployment in metro embedded platforms for estimating metro crowd density.


\begin{table*}[]
\centering
\caption{The inference speed of the MCNet and other competitors on the embedded device.}\label{Tbl7}
\setlength{\tabcolsep}{0.2em}
\begin{tabular}{clllllll}
\hline
\textbf{Models}                                               & \textbf{MCNet w/o}  & \textbf{MCNet w/}    & SqueezeNet              & MobileNet               & AlexNet                 & GoogleNet               & ResNet34                 \\ \hline
\begin{tabular}[c]{@{}l@{}}$\rm{Speed_{CPU}}$\\ (ms/frame)\end{tabular} & \multicolumn{1}{c}{26} & \multicolumn{1}{c}{30} & \multicolumn{1}{c}{58} & \multicolumn{1}{c}{102} & \multicolumn{1}{c}{170} & \multicolumn{1}{c}{223} & \multicolumn{1}{c}{453} \\ \hline
\end{tabular}
\end{table*}
Fig.~\ref{FIG:7} draws the precision recall curves of all networks on the SY\_METRO dataset, which can clearly demonstrates their recognition capability in three crowd density levels of this dataset. As is shown in Fig.~\ref{FIG:7}, the performance gap between MCNet and the other competitors is small, achieving a comparable metro crowd density recognition capability. The red lines of MCNet with IMA modules are higher than the light blue lines of the regular MCNet, indicating that the prediction accuracy of the plain network is improved at all crowd density levels.

Fig.~\ref{FIG:8} visualizes the intermediate feature representations of different layers of the MCNet with the IMA module that process images with different crowd density levels. In the bottom layer, the MCNet extracts low-level features and reinforces edge contours of passengers. In top layers, after convolution and pooling operations, the MCNet produces stronger feature activation in crowded passenger areas. Depending on the different crowd densities, it has different spatial distribution states on the feature map, showing that the high-density crowd has a broader feature activation, which gradually decreases as the crowd density level decreases, and the MCNet uses this property to recognize the density level of metro crowd image. It should be noted that the IMA module effectively improves the activation range and intensity of crowd texture features, making crowd texture features with different crowd density level more obvious, which helps the network to accurately classify the crowd images.

\begin{itemize}
  \item [1)]
  The MCNet takes only 26ms from reading an image to outputting the recognition result, and demonstrates a fast inference speed. This means that it can quickly deliver the density information of passengers in the carriage to the metro video surveillance system.
  \item [2)]
  After plugging the IMA module into the whole network, an additional 4ms inference time overhead is added, which makes the total inference time of MCNet reach 30ms. still maintains a faster inference speed. It is shown that the IMA module improves the network prediction accuracy with a limited increase in computational complexity, which is suitable for solving the problem of crowd density classification in embedded scenarios, and helps the plain network to capture crowd density information quickly and accurately.
\end{itemize}

\subsection{The experiments on the embedded device}
The experiments on the workstation have shown that the MCNet has the potential to be deployed on the embedded platforms. In order to accurately estimate its actual inference speed on the embedded device with limited hardware resources, this paper utilizes the C++ programming language write our model's forward inference program for metro crowd density estimation, and then test the inference speed of this program on the embedded device. The experimental results are shown in Table~\ref{Tbl7}, showing that:

\begin{table}[!h]
\centering
\caption{The energy consumption of all networks}\label{Tbl8}
\setlength{\tabcolsep}{0.2em}
\begin{tabular}{llll}
\hline
\multirow{2}{*}{\textbf{Models}} & \multicolumn{3}{c}{\textbf{Energy Consumption}} \\ \cline{2-4}
                                    & Max(W)          & Min(W)            & Ave(W)          \\ \hline
AlexNet                                 & 58           & 53 & 56.39            \\
GoogleNet                           & 58            & 53 & 56.38            \\
ResNet34                        & 58            & 53 & 56.46            \\
MobileNet                        & 58            & 53 & 56.44            \\
SqueezeNet                        & 61            & 50 & 56.25            \\
\textbf{MCNet w/o}                       & 60            & 52 & 55.42
\\
\textbf{MCNet w/}                       & 78            & 52 & 61.21        \\ \hline
\end{tabular}
\end{table}

\subsection{The power consumption experiments}
In embedded platforms, the power consumption of the model is a key performance metric, which saves energy cost of target platform and ensures the model runs for a long time in battery mode. In order to accurate evaluate the power usage of the proposed network, this paper measures the GPU power usage of the proposed network and other competitors in their inference stage with a interval of 0.01 seconds, and their detailed energy cost have been listed in Table~\ref{Tbl8}. As shown in the results of this Table, the MCNet only consumes 55.42W (average value) and demonstrates lower power consumption compared to other baseline networks, and the other competitors have higher power consumption (e.g. ResNet consumes 56.46W) and aren't friendly for deployment on the embedded systems with tight energy constraints compared to the proposed network. In particular, its KB-level total network parameters give it the potential to be stored directly in the edge device's on-chip memory, further reducing the power consumed by off-chip memory accesses in that device. However, once the IMA module is fitted to the MCNet, its operational power consumption fluctuates considerably, with the average power consumption rising to 61.21W, which is 5.79W higher than that of the plain network, suggesting that the IMA module increases the power consumption and affects its online operational performance in battery-powered embedded systems. The reason for this phenomenon is that the IMA module has multi-branch attentional convolutional operations and therefore consumes more power.



\section{Conclusion}\label{sec:4}
This paper has proposed a series of network structure and architecture designing strategies to construct the small network with competitive crowd density estimation performance and a novel IMA module to further promote the prediction accuracy of the plain network to classify crowd density level with the small computational burdens. This paper combines the IMA module with the newly designed small network to construct MCNet and validate it on various datasets. The ablation experiment results have demonstrated the generalisation ability of the IMA module to enhance the comprehensive prediction performance of the plain network in different scenarios, helping it deliver fast and accurate classification results with small extra computational cost overheads. The MCNet has obtained competitive crowd density estimation performance on benchmark datasets, especially on the SH\_METRO dataset, which has shown accurate crowd density classification capability as well as significant strengths in fewer total network parameters and faster inference speed, so it can be used to solve the metro passengers density estimation problem on the resource-constrained metro embedded platform. More importantly, the deployment experiments on the embedded device have once again supported the feasibility of applying the MCNet to metro embedded platform, thus meeting the application requirements of metro video surveillance and being used to construct a soft instrument to assist metro managers. In addition, the proposed MCNet can be utilized to solve other similar embedded vision problems with some modifications.

\noindent\textbf{Author Contributions} Qiang Guo: Writing-Original draft preparation, Carrying out the experiments, Methodology, Data curation Rubo Zhang: Supervision, Reviewing and Editing. Di Zhao: Contributing to experiments.
\newline

\noindent\textbf{Funding} No funding was received to carry out this study.
\newline

\noindent\textbf{Data availability and access} The benchmark datasets used in this paper come from
these papers \cite{Ref28,Ref24,Ref25,Ref26}. The SH\_METRO dataset are not available due to commercial restrictions.

\section*{Declarations}

\noindent\textbf{Competing Interests} The authors have no competing interests to disclose in any material discussed in this article.
\newline

\noindent\textbf{Ethical and informed consent for data used} Not applicable.
\newline

\bibliographystyle{splncs03}
\bibliography{example}


\end{document}